\documentclass[journal]{IEEEtran}

\usepackage{amsthm}
\usepackage{multirow}
\usepackage{makecell}

\newtheorem{assumption}{Assumption}

\newtheorem{theorem}{Theorem}
\newtheorem{corollary}[theorem]{Corollary}

\usepackage{times}
\usepackage{soul}
\usepackage{url}
\usepackage[utf8]{inputenc}
\usepackage{booktabs}

\usepackage{amsmath,amssymb,amsfonts}
\usepackage{algorithm}
\usepackage{algpseudocode}
\usepackage{amsmath}
\algnewcommand{\Phase}[1]{%
   \State \textbf{#1}%
}
\algnewcommand{\Var}[1]{\texttt{#1}}

\usepackage{graphicx}
\usepackage{textcomp}
\usepackage{xcolor}
\usepackage{multirow}
\usepackage{algorithm}
\usepackage{amsthm,amsmath,amssymb}
\usepackage{mathrsfs}
\usepackage{hyperref}
\usepackage{booktabs} % 用于 \toprule, \midrule, \bottomrule
\usepackage{siunitx}  % 用于 S 列格式
\usepackage{subfigure}
\renewcommand{\algorithmicrequire}{\textbf{Input:}}  % Use Input in the format of Algorithm
\renewcommand{\algorithmicensure}{\textbf{Output:}} % Use Output in the format of Algorithm
\def\BibTeX{{\rm B\kern-.05em{\sc i\kern-.025em b}\kern-.08em
    T\kern-.1667em\lower.7ex\hbox{E}\kern-.125emX}}
    
\hypersetup{colorlinks = true, %将链接文字带颜色
	    linkcolor = black, %图表引用颜色设置为蓝色
	    urlcolor = blue,%网页链接为蓝色
           citecolor = blue} %文献引用颜色设置为蓝色

\title{NCSAM: Noise-Compensated Sharpness-Aware Minimization for Noisy Label Learning}
\author{
Jiayu Xu, Jiaxin Deng, and Junbiao Pang
\thanks{Jiayu Xu and Junbiao Pang are with Beijing University of Technology.}
\thanks{Corresponding author: Junbiao Pang (e-mail: junbiao\_pang@bjut.edu.cn).}
}
\begin{document}
\maketitle

\begin{abstract}
%已改
% Learning from Noisy Labels (LNL) presents a fundamental challenge in deep learning, as real-world datasets often contain corrupted annotations. Current research focuses on sophisticated noisy label correction mechanisms. In contrast, this paper establishes a theoretical analysis of the relationship between the flatness of the loss landscape and noisy labels. Concretely, we propose Noise-Compensated Sharpness-Aware Minimization (NCSAM) to leverage the noise-compensated perturbation effectively to mitigate the damage from noisy labels and further improve the model's generalization performance. The behavior of NCSAM on noisy label datasets breaks the memory effect of noisy labels.
% Extensive experimental results on multiple benchmark datasets demonstrate the consistent superiority of the proposed method over existing state-of-the-art approaches on diverse tasks. The code is publicly available at:~\url{}.
Learning from Noisy Labels (LNL) remains a fundamental challenge in deep learning because real-world datasets often contain corrupted annotations. Most existing methods rely on label correction or sample selection mechanisms. In contrast, we study LNL from an optimization perspective by establishing a theoretical connection between label noise and the flatness-seeking behavior of Sharpness-Aware Minimization (SAM). Based on this analysis, we propose Noise-Compensated Sharpness-Aware Minimization (NCSAM), which uses a noise-compensated perturbation to counteract the optimization bias induced by noisy labels. By correcting distorted SAM perturbations, NCSAM mitigates the memorization of noisy labels during training while preserving the simplicity of optimization-based learning. Experiments on synthetic and real-world noisy-label benchmarks show that NCSAM consistently improves over SAM-based optimization baselines and remains competitive with representative noisy-label learning methods.

\end{abstract}


% 后面用AI改，提示词，修改下面的语句，让其更符合顶级科技期刊（如TPAMI，IJCV）的写作风格,逻辑，并更正其中的语法错误，让文字简洁表达，但不改变其中的latex语法和引用，并给出详细修改原因

%以按你的提示词更改
\section{Introduction}\label{sec:intro}

Learning from Noisy Labels (LNL) is an important yet challenging problem in deep learning. In practice, large-scale datasets are often collected through crowdsourcing, where annotation errors are inevitable~\cite{song-2019-selfie}. The biased gradients induced by noisy labels corrupt the supervision signal and can severely degrade model generalization.

Most existing approaches to LNL aim to reduce the adverse effects of noisy labels during deep neural network (DNN) training. These methods can be broadly grouped into three categories: 1) sample selection methods exploit the memorization effect of deep networks and use early-stage models to identify likely clean samples or refine the labels of selected instances~\cite{cheng-2021-Learning}\cite{karim-2022-unicon}; 2) label correction methods either explicitly rectify noisy labels~\cite{tanaka-2018-joint} or replace them with pseudo-labels~\cite{li-2021-learning}\cite{li-2023-disc}; and 3) regularization methods adopt robust loss functions~\cite{englesson-2021-generalized}\cite{zhou-2023-asymmetric} or strong data augmentation strategies (\textit{e.g.}, MixUp~\cite{zhang-2018-mixup}) to smooth decision boundaries and reduce the influence of mislabeled samples~\cite{englesson-2021-generalized}\cite{sarfraz-2021-noisy}.
However, these approaches still face fundamental limitations. Sample selection can become unreliable when the model is overconfident, causing mislabeled samples to remain in the presumed clean set. Label correction and regularization methods, in turn, still depend heavily on the model's own generalization ability to produce reliable pseudo-labels or robust decision boundaries. Ultimately, the core difficulty of LNL lies in the biased gradients introduced by noisy labels, which drive optimization toward suboptimal solutions and degrade generalization.

In this paper, we address LNL from an optimization perspective, motivated by theoretical insights from the PAC-Bayesian framework. Specifically, we show that noisy labels distort the flatness-seeking behavior of Sharpness-Aware Minimization (SAM), which motivates our Noise-Compensated Sharpness-Aware Minimization (NCSAM). Unlike prior methods that rely on complex label correction~\cite{zheng-2021-meta} or heuristic early-learning schedules~\cite{liu-2020-early}, NCSAM provides a general optimization strategy that can be combined with arbitrary loss functions under noisy supervision. The main contributions of this work are summarized as follows:
\begin{itemize}
\item Our theoretical analysis shows that label noise yields a looser generalization bound than clean supervision. It also explains why existing SAM-based methods become suboptimal under noisy labels, consistent with the empirical observations in~\cite{deng-2024-bsam}.

\item We introduce a noise-compensated mechanism that explicitly aligns parameter perturbations with noise-induced parameter deviations. This design mitigates overfitting to noisy labels and improves the robustness of SAM-style optimization without relying on sample selection or label correction.
\end{itemize}

\section{Related Work}
\subsection{Learning from Noisy Labels}  
%缩减缩减

\textbf{Noise-Robust Loss Functions.} One line of research designs loss functions with intrinsic robustness to noisy labels. For example, Generalized Cross Entropy (GCE)~\cite{zhang-2018-generalized} trades off Mean Absolute Error (MAE) and Cross Entropy (CE), where MAE is known to be more robust to label noise. Symmetric Cross Entropy (SCE)~\cite{wang-2019-symmetric} further introduces asymmetric penalties to reduce the disturbance caused by noisy labels. Although such approaches offer theoretical robustness, their practical performance often depends heavily on careful hyper-parameter tuning. Importantly, these methods are largely orthogonal to our optimization-based approach.

\textbf{Cleaning Up Noisy Labels.} Another line of work alleviates noisy labels by identifying, correcting, or down-weighting corrupted samples. For instance, curriculum-based strategies such as Dynamic Instance Hardness (DIH)~\cite{zhou-2020-curriculum} use an exponential moving average of model predictions to suppress unreliable samples. Data augmentation techniques such as Mixup~\cite{zhang-2017-mixup} reduce the tendency of deep networks to memorize noisy annotations. Label correction methods~\cite{chen-2021-beyond}\cite{huang-2022-balance}\cite{zheng-2020-error} refine noisy labels using the model's own predictions, but this self-dependency limits their reliability, especially under high noise rates. Alternatively, Co-teaching~\cite{han-2018-co} employs mutual supervision between peer networks to filter noisy samples. Despite their effectiveness, overly aggressive sample removal or correction may cause the model to overfit the remaining clean subset, thereby limiting generalization.

\textbf{Alternative Paradigms for Noisy Label Learning.}  
Several studies reformulate LNL from alternative perspectives. Specifically, semi-supervised learning methods treat confidently predicted samples as labeled data and uncertain samples as unlabeled data~\cite{laine-2016-temporal}\cite{tarvainen-2017-mean}. Meta-learning approaches~\cite{ren-2018-learning}\cite{shu-2019-meta} use a small clean validation set to guide optimization under noisy supervision. Unsupervised and contrastive learning methods~\cite{huang-2022-contrastive}\cite{xu-2023-label} aim to learn representations that are more robust to label corruption. More recently, hybrid frameworks that combine multiple perspectives have been proposed~\cite{li-2020-dividemix}\cite{huang-2023-combining} to jointly reweight samples and improve representation learning. %However, their designs are often driven by heuristic principles rather than explicitly modeling generalization behavior in the parameter space. 
To the best of our knowledge, our method is orthogonal to these approaches and is among the first attempts to address noisy labels through sharpness-aware optimization.

\subsection{Sharpness and Noisy Labels}
Sharpness has been widely studied as a quantitative measure of generalization~\cite{foret-2020-sharpness}. For example, He et al.~\cite{he-2019-asymmetric} proposed trajectory averaging strategies to locate flat minima that exhibit improved robustness. In recent years, SAM has emerged as a powerful family of optimization techniques that explicitly seek flat minima. Although effective across various tasks, their performance can still degrade in more challenging scenarios. For example, prior work has explored reweighting strategies to adapt SAM to data imbalance or long-tailed settings~\cite{liu-2025-balanced}\cite{rangwani-2022-escaping}, but the perturbation design remains largely heuristic and lacks explicit consideration of noisy labels. In contrast, our NCSAM extends SAM by introducing a theoretically motivated noise-compensated perturbation mechanism specifically desMigned to improve generalization under noisy labels.

% Recently, several studies have attempted to adapt sharpness-aware optimization to noisy-label learning. For example, Clean-aware SAM~\cite{huang-2025-learning} estimates clean subsets and constructs perturbations primarily based on clean samples to improve optimization stability under noisy supervision. In contrast, our method does not rely on explicit clean/noisy sample partitioning or sample selection. Instead, NCSAM models noisy supervision from a perturbation-bias perspective and directly compensates for the dominant noise-induced deviation in SAM perturbations. Therefore, our method addresses noisy-label learning through perturbation compensation rather than clean-sample estimation, making it fundamentally different from existing clean-aware sharpness-aware optimization approaches.
Recent noisy-label variants of SAM often follow a clean-aware strategy. For example, Clean-aware SAM~\cite{huang-2025-learning} estimates clean subsets and constructs perturbations mainly from clean samples. By contrast, NCSAM does not partition or select samples. It treats label noise as a bias in the SAM perturbation and compensates this bias directly, addressing noisy-label learning through perturbation correction rather than clean-sample estimation.

\section{Method}
\subsection{The Generation of Noisy Labels} 
\label{The Generation of Noisy Labels}

Let the input space be $\mathcal{X}$ and the label space be $\mathcal{Y}=\{1,\ldots,C\}$, where $C$ denotes the number of classes. Given a label-noisy training set
$\tilde{\mathcal{D}}=\{(x_i,\tilde{y}_i)\}_{i=1}^{n}$,
where $x_i \in \mathcal{X}$ denotes an input sample and $\tilde{y}_i \in \mathcal{Y}$ is its observed label.
Unlike clean supervision, the observed label $\tilde{y}_i$ may differ from the latent clean label $y_i$.

We assume that each observed label $\tilde{y}_i$ is corrupted with an instance-specific probability $p_i$. The noisy labeling process is defined as :
\begin{equation}\label{eq:noise_label_mech}
\tilde{y}_{i}=\left\{\begin{array}{ll}
y_i, & \text{with probability } 1-p_i,\\
u_i, & \text{with probability } p_i,
\end{array}\right.
\end{equation}
where $u_i\sim \mathcal{T}(\cdot\mid x_i,y_i)$ and $u_i\ne y_i$ denotes a corrupted label sampled from a label-transition mechanism $\mathcal{T}$. Eq.~\eqref{eq:noise_label_mech} models instance-dependent noise, where the sample-wise corruption probability is drawn from a Beta distribution:
\begin{equation}
p_i \sim \mathrm{Beta}(\theta_1,\theta_2),
\end{equation}
where $\theta_1$ and $\theta_2$ are positive shape parameters. We use the Beta distribution because the corruption probability is naturally bounded in $[0,1]$, and the two-parameter family can represent clean-dominated, noise-dominated, and highly heterogeneous corruption patterns. Specifically, $\mathbb{E}[p_i]=\theta_1/(\theta_1+\theta_2)$; hence $\theta_1\ll\theta_2$ corresponds to mostly clean samples, whereas $\theta_1\gg\theta_2$ corresponds to a heavily corrupted training set. When $\theta_1$ and $\theta_2$ are comparable, the distribution assigns substantial mass to intermediate corruption probabilities, which captures ambiguous or hard samples. This assumption is used as a flexible probabilistic model for instance-dependent noise rather than as a requirement that real-world label noise exactly follows a Beta law. NCSAM does not require estimating $\theta_1$ or $\theta_2$; the model only motivates the heterogeneity of instance-dependent corruption.

Given a parametric model $f(x;\mathbf{w})$ with parameters $\mathbf{w}\in\mathbb{R}^{K}$ and loss function $L(\mathbf{w};x,y)$, for any mini-batch $\mathcal{B}$, we denote its clean and noisy parts by $\mathcal{S}_{\mathcal{B}}$ and $\hat{\mathcal{S}}_{\mathcal{B}}$, respectively, satisfying $\mathcal{B}=\mathcal{S}_{\mathcal{B}}\cup\hat{\mathcal{S}}_{\mathcal{B}}$. Let $\mathbf{g}_{\mathrm{clean}}$ be the average gradient over $\mathcal{S}_{\mathcal{B}}$ and $\mathbf{g}_{\mathrm{noise}}$ be the average gradient over $\hat{\mathcal{S}}_{\mathcal{B}}$. The mini-batch gradient can be decomposed as:
\begin{equation}
\mathbf{g}_{\mathcal{B}}
=
\frac{|\mathcal{S}_{\mathcal{B}}|}{|\mathcal{B}|}\mathbf{g}_{\mathrm{clean}}
+
\frac{|\hat{\mathcal{S}}_{\mathcal{B}}|}{|\mathcal{B}|}\mathbf{g}_{\mathrm{noise}}.
\end{equation}
The noisy component induces a parameter deviation $\Delta\mathbf{w}_{\mathrm{noise}}= -\eta \mathbf{g}_{\mathrm{noise}}$, leading to biased parameters $\tilde{\mathbf{w}} = \mathbf{w} + \Delta\mathbf{w}_{\mathrm{noise}}$. This noise-induced deviation biases optimization and ultimately harms generalization. Our question is therefore: \textit{can we counteract the parameter deviation caused by noisy labels without explicitly correcting the labels themselves?}

\subsection{PAC-Bayes Motivated Analysis of SAM under Noisy Labels}
\label{sec:theory}
We analyze the effect of noisy labels through an oracle decomposition of the training set. Specifically, let $\mathcal{S}$ denote the subset whose observed labels coincide with the latent clean labels, and let $\hat{\mathcal{S}}$ denote the subset with corrupted labels. This decomposition is introduced only for theoretical analysis and is not required by the proposed algorithm. In practice, training is performed on the mixed set $\mathcal{M}=\mathcal{S}\cup\hat{\mathcal{S}}$, where $\mathcal{S}$ contains $n_s$ clean-label samples and $\hat{\mathcal{S}}$ contains $n_n$ noisy-label samples. Let $n=n_s+n_n$ and $\alpha=n_n/n$ , where $\alpha$ denote the realized noisy-sample fraction. The mixed empirical risk optimized by training is:
\begin{equation}
L_{\mathcal{M}}(\mathbf{w})=\frac{1}{n}\left(n_sL_{\mathcal{S}}(\mathbf{w})+n_nL_{\hat{\mathcal{S}}}(\mathbf{w})\right).
\label{eq:mixed_risk}
\end{equation}
Since the target distribution is the clean data distribution $\mathcal{D}$, the PAC-Bayesian empirical term must first be written on the clean subset $\mathcal{S}$. The mixed risk is then introduced through the exact identity in Eq.~\eqref{eq:mixed_risk}. Following the perturbation-based PAC-Bayesian derivation in SAM Appendix A.1~\cite{foret-2020-sharpness}, we obtain the following noisy-label counterpart.

\begin{assumption}
\label{assump:theory}
The loss is bounded, i.e., $0\le L(\mathbf{w};x,y)\le1$. The clean subset $\mathcal{S}$ is sampled from the target distribution $\mathcal{D}$, while the clean/noisy decomposition is used only for analysis. The PAC-Bayesian prior is independent of $\mathcal{S}$, and the SAM perturbation condition
$L_{\mathcal{D}}(\mathbf{w})\le \mathbb{E}_{\boldsymbol{\epsilon}\sim\mathcal{N}(\mathbf{0},\sigma^2\mathbf{I}_K)}[L_{\mathcal{D}}(\mathbf{w}+\boldsymbol{\epsilon})]$
holds for the considered solution, where $\mathbf{I}_K$ denotes the $K$-dimensional identity matrix.
\end{assumption}

\begin{theorem}[Generalization Bound under Noisy Labels]
\label{thm:noisy_sam_bound}
Under Assumption~\ref{assump:theory}, suppose $n_s=n(1-\alpha)>1$. For any $\rho>0$, set:
\begin{equation}
\sigma=
\frac{\rho}{
\sqrt{K}\left(1+\sqrt{\frac{\log n_s}{K}}\right)
}.
\end{equation}
Then, conditioning on the noisy subset $\hat{\mathcal{S}}$, with probability at least $1-\delta$ over the clean sample draw $\mathcal{S}$, the following bound holds:
\begin{equation}
\begin{split}
L_{\mathcal{D}}(\mathbf{w})
\leq& \frac{1}{1-\alpha}\max_{\|\boldsymbol{\epsilon}\|_2\le\rho}\!L_{\mathcal{M}}(\mathbf{w}+\boldsymbol{\epsilon}) \\
&+ \frac{1}{(1-\alpha)\sqrt{n_s}} + \sqrt{\frac{\mathcal{X}}{n_s-1}},\\
\mathcal{X}=&\frac14K\log\left(\!1+\frac{\|\mathbf{w}\|_2^2}{\rho^2}\left(1+\sqrt{\frac{\log n_s}{K}}\right)^{\!2}\right)+\frac14 \\
&+\log\frac{n_s}{\delta}+2\log(6n_s+3K).
\end{split}
\label{eq:noisy_sam_bound_alpha}
\end{equation}
 where $K$ is the number of model parameters and $\rho$ is the SAM neighborhood radius. The posterior may be data-dependent, while the prior is chosen independently of the clean sample as in the standard PAC-Bayesian setting.
\end{theorem}

\begin{corollary}[Unconditional interpretation]
\label{cor:unconditional_bound}
If the noisy subset $\hat{\mathcal{S}}$ is also generated randomly and the conditions of Theorem~\ref{thm:noisy_sam_bound} hold for every realized $\hat{\mathcal{S}}$ with fixed $n_s$ and $n_n$, then Eq.~\eqref{eq:noisy_sam_bound_alpha} also holds with probability at least $1-\delta$ over the joint draw of $(\mathcal{S},\hat{\mathcal{S}})$. This follows directly from the law of total probability, since
\[
\Pr(E)=\mathbb{E}_{\hat{\mathcal{S}}}\!\left[\Pr(E\mid\hat{\mathcal{S}})\right]\ge 1-\delta,
\]
where $E$ denotes the event that Eq.~\eqref{eq:noisy_sam_bound_alpha} holds. When the number of clean samples is random, the statement should be interpreted conditional on the realized value of $n_s$, or alternatively with a standard union bound over possible values of $n_s$.
\end{corollary}

\textbf{Derivation sketch.}
The complete derivation is provided in Appendix~\ref{app:noisy_sam_bound}. The key step is to start from the PAC-Bayesian bound on the clean subset $\mathcal{S}$ and then introduce the mixed empirical risk through the exact identity:
\begin{equation}
L_{\mathcal{S}}(\mathbf{w}+\boldsymbol{\epsilon})
=
\frac{n}{n_s}L_{\mathcal{M}}(\mathbf{w}+\boldsymbol{\epsilon})
-
\frac{n_n}{n_s}L_{\hat{\mathcal{S}}}(\mathbf{w}+\boldsymbol{\epsilon})
\le
\frac{1}{1-\alpha}L_{\mathcal{M}}(\mathbf{w}+\boldsymbol{\epsilon}),
\label{eq:clean_to_mixed_bound}
\end{equation}
where the inequality follows from $L_{\hat{\mathcal{S}}}(\mathbf{w}+\boldsymbol{\epsilon})\ge0$. Substituting Eq.~\eqref{eq:clean_to_mixed_bound} into the PAC-Bayesian bound used by SAM and then applying the Gaussian posterior, prior-variance grid, union-bound, and Gaussian-tail steps of SAM Appendix A.1 yields Eq.~\eqref{eq:noisy_sam_bound_alpha}. The only additional factor introduced by label noise is the clean-to-mixed conversion factor $n/n_s=1/(1-\alpha)$.

\textbf{Remark.}
Eq.~\eqref{eq:noisy_sam_bound_alpha} shows that label noise weakens the SAM-style guarantee in two ways. First, the mixed neighborhood risk is amplified by $1/(1-\alpha)$, meaning that the same value of $\max_{\|\boldsymbol{\epsilon}\|_2\le\rho}L_{\mathcal{M}}(\mathbf{w}+\boldsymbol{\epsilon})$ leads to a looser clean-distribution bound when the clean fraction decreases. Second, the complexity term depends on the effective clean sample size $n_s=n(1-\alpha)$. This analysis does not introduce an additional $\alpha/(1-\alpha)$ penalty; such a term does not follow from the required upper-bound direction.

\textbf{Comparison with clean SAM.}
When $\alpha=0$, Eq.~\eqref{eq:noisy_sam_bound_alpha} reduces to the standard SAM-style PAC-Bayesian form derived from Appendix A.1 of SAM~\cite{foret-2020-sharpness}. For $\alpha>0$, the leading empirical term becomes:
\begin{equation}
\frac{1}{1-\alpha}
\max_{\|\boldsymbol{\epsilon}\|_2\le\rho}L_{\mathcal{M}}(\mathbf{w}+\boldsymbol{\epsilon}),
\label{eq:noise_amplification_factor}
\end{equation}
whereas the clean SAM counterpart has no amplification factor. Therefore, even if the mixed neighborhood risk has the same numerical value as the clean neighborhood risk, the noisy-label bound is enlarged by $1/(1-\alpha)$ and its complexity term is evaluated with the smaller effective sample size $n_s=n(1-\alpha)$.
For example, $\alpha=0.4$ yields an amplification factor of $1.67$, while $\alpha=0.6$ yields $2.50$, which makes the weakening effect of label noise explicit.

More concretely, label noise changes the perturbation direction used by SAM because the ascent direction is computed from mixed clean and noisy gradients.  For a mini-batch $\mathcal{B}$, let $\lambda_{\mathcal{B}}=|\hat{\mathcal{S}}_{\mathcal{B}}|/|\mathcal{B}|$ denote its noisy-sample proportion. The mixed gradient used by standard SAM can be written as:
\begin{equation}
\boldsymbol{\epsilon}_{\mathrm{SAM}}
=
\rho\frac{\mathbf{g}_{\mathcal{B}}}{\|\mathbf{g}_{\mathcal{B}}\|}
=
\rho
\frac{(1-\lambda_{\mathcal{B}})\mathbf{g}_{\mathrm{clean}}+\lambda_{\mathcal{B}}\mathbf{g}_{\mathrm{noise}}}
{\|(1-\lambda_{\mathcal{B}})\mathbf{g}_{\mathrm{clean}}+\lambda_{\mathcal{B}}\mathbf{g}_{\mathrm{noise}}\|},
\label{eq:sam_perturbation}
\end{equation}
where $\lambda_{\mathcal{B}}$ equals $\alpha$ only in expectation or when the mini-batch matches the global noise proportion. For mislabeled samples, $\mathbf{g}_{\mathrm{noise}}$ often points in the opposite direction of $\mathbf{g}_{\mathrm{clean}}$ (i.e., $\mathbb{E}[\langle \mathbf{g}_{\mathrm{clean}}, \mathbf{g}_{\mathrm{noise}} \rangle] < 0$), which rotates the perturbation direction away from the steepest ascent direction of the clean loss landscape. This misalignment means that SAM is no longer searching for flat minima in the clean loss landscape, but rather in the distorted noisy loss landscape.

The bound and the perturbation decomposition together motivate NCSAM. The bound shows that the clean-distribution guarantee is controlled by the mixed neighborhood risk, which is amplified as the clean fraction decreases. The perturbation decomposition further indicates that this degradation is operationally reflected in the biased SAM perturbation computed from mixed gradients. Since the true deviation $\Delta\mathbf{w}_{\mathrm{noise}}$ is not observable, NCSAM estimates its dominant direction through temporary label flipping and uses the estimate only to compensate the perturbation, rather than to relabel or discard training samples.

\subsection{Noise-Compensated SAM}\label{sec:ncsam}

Following the Gaussian perturbation model used in Theorem~\ref{thm:noisy_sam_bound}, we estimate and compensate for the noise-induced parameter deviation while minimizing the smoothed noisy loss $\mathbb{E}_{\boldsymbol{\epsilon} \sim \mathcal{N}(\mathbf{0},\sigma^2 \mathbf{I}_K)}\left[ L_{\hat{\mathcal{S}}}(\mathbf{w} + \Delta\mathbf{w}_{\mathrm{noise}} + \boldsymbol{\epsilon}) \right]$. Concretely, let $\mathbf{g}^*_{\mathrm{noise}}$ denote the gradient computed from temporarily label-flipped samples. We define the compensation term as:

\begin{equation}\label{eq:dw}
\Delta\mathbf{w}_c = -s(t)\mathbf{g}^*_{\mathrm{noise}},
\end{equation}
where $s(t)$ is a time-dependent scaling function that increases during the early stage of training and saturates after a prescribed number of epochs.

To remove the influence of noisy labels from the SAM perturbation $\tilde{\boldsymbol{\epsilon}}$:
\begin{equation}
\tilde{\boldsymbol{\epsilon}} = \Delta\mathbf{w}_{\mathrm{noise}} + \boldsymbol{\epsilon},
\label{eq:epsi}
\end{equation}
where $\Delta\mathbf{w}_{\mathrm{noise}}$ represents the noise-induced parameter deviation and $\boldsymbol{\epsilon}$ is the clean SAM perturbation. We then incorporate the correction term and obtain the adjusted perturbation:
$\boldsymbol{\epsilon}' = \tilde{\boldsymbol{\epsilon}} - \Delta\mathbf{w}_c. $ 
In implementation, the exact noise-induced deviation $\Delta\mathbf{w}_{\mathrm{noise}}$ is not observable. We therefore use the standard SAM perturbation $\boldsymbol{\epsilon}_{\mathrm{SAM}}$ computed from the mixed mini-batch gradient as the practical realization of the noisy perturbation direction, and subtract the estimated compensation term $\Delta\mathbf{w}_c$. The corrected perturbation is then projected back to the $\ell_2$ ball with radius $\rho$ to remain consistent with the SAM neighborhood constraint:

\begin{equation}
\boldsymbol{\epsilon}_{\mathrm{NCSAM}}
=
\operatorname{Proj}_{\|\boldsymbol{\epsilon}\|_2\le\rho}
\left(
\boldsymbol{\epsilon}_{\mathrm{SAM}}-\Delta\mathbf{w}_c
\right).
\label{eq:ncsam_perturbation}
\end{equation}
% This adaptive adjustment compensates for the noise-induced bias in the perturbation and stabilizes sharpness-aware optimization under noisy supervision.
Here $\operatorname{Proj}_{\|\boldsymbol{\epsilon}\|_2\le\rho}(\cdot)$ denotes the Euclidean projection onto the $\ell_2$ ball of radius $\rho$:
\begin{equation}
\operatorname{Proj}_{\|\boldsymbol{\epsilon}\|_2\le\rho}(\mathbf{v})
=
\arg\min_{\|\boldsymbol{\epsilon}\|_2\le\rho}\|\boldsymbol{\epsilon}-\mathbf{v}\|_2.
\end{equation}
Thus, $\boldsymbol{\epsilon}_{\mathrm{NCSAM}}$ is the projected feasible form of the compensated perturbation $\boldsymbol{\epsilon}'=\boldsymbol{\epsilon}_{\mathrm{SAM}}-\Delta\mathbf{w}_c$, and we use it in the subsequent neighborhood maximization. This adaptive adjustment compensates for the noise-induced bias in the perturbation and stabilizes sharpness-aware optimization under noisy supervision.

For a mini-batch $\mathcal{B}$, let
$L_{\mathcal{B}}(\mathbf{w})=|\mathcal{B}|^{-1}\sum_{(x_i,\tilde{y}_i)\in\mathcal{B}}L(\mathbf{w};x_i,\tilde{y}_i)$.
Accordingly, the Noise-Compensated SAM (NCSAM) objective becomes:
\begin{equation}\label{eq:sam}
\min_{\mathbf{w}} \max_{\|\boldsymbol{\epsilon}'\|\le\rho} L_{\mathcal{B}}(\mathbf{w} + \boldsymbol{\epsilon}'), 
\end{equation}

In summary, the corrected perturbation $\boldsymbol{\epsilon}'$ is designed to mitigate the distortion introduced by noisy gradients in the standard SAM update. 
%As shown in Section~\ref{PAC-Bayes Motivated Analysis of SAM under Noisy Labels}, label noise induces a systematic bias that contaminates both the direction and magnitude of the SAM perturbation through the accumulated deviation $\Delta \mathbf{w}$. 
Rather than assuming the noise-induced deviation can be completely eliminated, NCSAM explicitly compensates for its dominant component by introducing the correction term $\Delta\mathbf{w}_c$.

\subsubsection{Dynamic Compensation Scaling}
\label{sec:dynamic_compensation_schedule}

The compensation strength should not be fixed throughout training. At the early stage, the network predictions are unstable, and an aggressively estimated noise direction may disturb the learning of clean discriminative patterns. As training proceeds, the model becomes more reliable for identifying structured noisy directions, and a stronger compensation is beneficial for suppressing the bias induced by corrupted labels. To this end, we adopt a bounded dynamic schedule for the compensation coefficient.

Let $t\in[0,1]$ denote the normalized training progress. The scaling coefficient is defined as:
\begin{equation}
s(t)=
\left\{
\begin{array}{ll}
\kappa r(t), & r(t)<1,\\
\kappa, & r(t)\geq 1,
\end{array}
\right.
\label{eq:dynamic_schedule}
\end{equation}
where $r(t)=2(3-2t)t^2$ is a smooth increasing function and $\kappa$ is the maximum compensation strength. This schedule gradually activates the correction term $\Delta\mathbf{w}_c$ while keeping its magnitude bounded. The increasing phase reduces unreliable early compensation, whereas the saturation phase prevents over-compensation and preserves useful clean-gradient information during convergence.

\subsubsection{Noise-Gradient Estimation via Progressive Label Flipping}
\label{sec:noise_gradient_estimation}
%\subsubsection{Noise label Simulation}

In this work, we approximate $\mathbf{g}^*_{\mathrm{noise}}$ in Eq.~\eqref{eq:dw} through progressive label flipping based on model logits. The idea is to identify samples that are more likely to be mislabeled and temporarily flip them to semantically similar categories. This raises two practical questions:
\begin{itemize}
    \item \textit{Which samples are likely to have noisy labels?}  
    An intuitive solution is to identify samples whose prediction confidence is low or whose logits exhibit small margins between the top predicted classes. These samples are more likely to be mislabeled since the model remains uncertain about their true category.

    \item \textit{How to flip their labels into reasonable noisy labels?}  
    An intuitive solution is to flip their labels to classes that lie near the decision boundary, such as the second-highest logit class or among the top-$k$ most probable classes. These classes are semantically closer to the model's belief, producing more realistic and structured noisy labels.
\end{itemize}

Specifically, given the output logits $\mathbf{z}_i \in \mathbb{R}^{C\times 1}$ for each sample $x_i$, we compute the gap between the top-2 logits:
\begin{equation}\label{eq:delta_i}
\delta_i = z_i^{(1)} - z_i^{(2)},
\end{equation}
where $z_i^{(1)}$ and $z_i^{(2)}$ denote the largest and second-largest logits, respectively. A smaller $\delta_i$ implies that the model is less certain about the classification decision and that the corresponding sample lies closer to the decision boundary. Therefore, samples with smaller $\delta_i$ are more likely to be affected by label noise.

We then design a probabilistic sampling mechanism based on the inverse of the logit gap $q_i$ as follows:

\begin{equation}\label{eq:pi}
p_i = \frac{q_i}{\sum_{j=1}^{|\mathcal{B}|} q_j}, \quad q_i = \frac{1}{1+\delta_i},
\end{equation}
% where $\frac{1}{1+\delta_i}$ makes $q_i$ be a long-tailed distribution which help every sample has certain probability to be selected as noisy labels. Because in real application, we barely distinguish which one sample defintely be immunitly corrupted into noisy labels.
where $\frac{1}{1+\delta_i}$ induces a long-tailed distribution over $q_i$, so even high-confidence samples retain a nonzero probability of being selected. This is desirable because, in practice, it is difficult to determine with certainty which samples are truly corrupted.
Given a label-flip ratio $r_{\mathrm{flip}}\in[0,1]$, we sample $m_{\mathrm{flip}}=\lfloor r_{\mathrm{flip}}|\mathcal{B}|\rfloor$ candidate samples from $\mathcal{B}$ according to the probabilities $\{p_i\}_{i=1}^{|\mathcal{B}|}$.

Once the subset of candidate samples is selected, we generate their temporary noisy labels by flipping each sample's label to the highest-logit class different from its observed label. This choice ensures that the flipped labels correspond to semantically similar categories, thereby mimicking asymmetric label errors that often appear in real-world noisy datasets, e.g., "dog" mislabeled as "wolf" rather than "truck".

Formally, for each selected sample $x_i$, its simulated noisy label $\hat{y}^*_i$ is defined as:
\begin{equation}\label{flip-2}
\hat{y}^*_i = \arg\max_{c \neq \tilde{y}_i} z_i^{(c)}.
\end{equation}
The resulting set of noisy samples $\{(x_i, \hat{y}^*_i)\}$ is then used to compute the simulated \textit{noise-induced gradient} $\mathbf{g}^*_{\mathrm{noise}}$, from which the weight offset $\Delta\mathbf{w}_c$ is estimated.

This strategy aims to align the artificially induced noise direction with the realistic noise-gradient distribution, providing a controllable and semantically consistent way to approximate noise-induced perturbations under real-world noisy conditions.

\textbf{Relationship with Label Correction.}

% Some existing studies attempt to extract useful supervisory signals from noisy data by generating pseudo-labels within semi-supervised learning frameworks. These methods typically rely on a label-correlation assumption to distinguish clean from noisy samples—a problem that remains challenging and largely unresolved. In contrast, our noisy-label simulation directly supplies the noise-compensation term $ \Delta \mathbf{w}_c $ required by NCSAM without explicitly identifying which samples are mislabeled. Thus, the two approaches are orthogonal and serve different purposes. Experimental results further demonstrate that our method can substantially improve the performance and generalization of existing label-correction-based approaches
Some existing studies attempt to extract useful supervisory signals from noisy data by generating pseudo-labels within semi-supervised learning frameworks. These methods typically rely on a label-correlation assumption to distinguish clean from noisy samples, which remains challenging under heavy noise. By contrast, our noisy-label simulation is not designed to recover clean annotations; instead, it estimates the compensation term $\Delta\mathbf{w}_c$ required by NCSAM. The two ideas are therefore orthogonal. Theoretically, our method can also complement label-correction approaches by improving their optimization robustness.

% \subsubsection{how to design $s(t)$}

% \begin{small}
% \begin{algorithm}[!t]
%     \renewcommand{\algorithmicrequire}{\textbf{Input:}}
%     \renewcommand{\algorithmicensure}{\textbf{Output:}}  
%     \caption{NCSAM Training Algorithm}
%     \label{alg:ncsam}
%     \begin{algorithmic}[1]
%         \Require Training set $\mathcal{D}_{train}$, model $f_W$; learning rate $\eta$; epochs $T$, warm-up $T_w$; SAM radius $\rho$; label-flip ratio $\gamma$,scaling function $s(t)$.  
        
%         \State Initialize model $\mathbf{w}$, optimizer (SGD/SAM), and scheduler.
%         \For{$t = 1$ to $T$}
%             \If{$t < T_w$}  
%                 \State Using optimizer SGD or SAM.
%                 \State Update $\mathbf{w}_{t+1} \leftarrow \mathbf{w}_{t} - \eta \nabla_\mathbf{w}\mathbf{L}(f_\mathbf{w}(x_i), \tilde{y}_i)$ .
%             \Else 
%                 \State Select uncertain samples by Eq.\ref{eq:delta_i} and Eq.\ref{eq:pi};
%                 \State Simulated label noise Eq.\ref{flip-2};
%                 \State Compute correction Eq.\ref{eq:dw}.
%                 \State SAM first step is to obtain a noisy perturbation;
%                 \State Add the correction parameters generated by Eq.\ref{eq:dw} to Eq.\ref{eq:epsi}; 
%                 \State SAM second step is to optimize Eq.\ref{eq:sam}.
%             \EndIf
            
%         \EndFor
%     \end{algorithmic}
% \end{algorithm}
% \end{small}
\begin{small}
\begin{algorithm}[!t]
    \renewcommand{\algorithmicrequire}{\textbf{Input:}}
    \renewcommand{\algorithmicensure}{\textbf{Output:}}  
    \caption{NCSAM Training Algorithm}
    \label{alg:ncsam}
    \begin{algorithmic}[1]
\Require Training set $\tilde{\mathcal{D}}$, model $f_{\mathbf{w}}$, learning rate $\eta$, epochs $T$, warm-up epochs $T_w$, SAM radius $\rho$, label-flip ratio $r_{\mathrm{flip}}$, and compensation schedule $s(\cdot)$.
\Ensure Trained parameters $\mathbf{w}$.
        \State Initialize $\mathbf{w}$ and optimizer.
        \For{$e=1$ to $T$}
            \For{each mini-batch $\mathcal{B}\subset\tilde{\mathcal{D}}$}
                \If{$e\le T_w$}
                    \State Update $\mathbf{w}$ by SGD on $\mathcal{B}$.
                \Else
                    \State Estimate $\mathbf{g}^*_{\mathrm{noise}}$  with Eqs.~\eqref{eq:delta_i}, \eqref{eq:pi}, and \eqref{flip-2}.
                    \State Compute $s(e/T)$ and $\Delta\mathbf{w}_c$ using Eqs.~\eqref{eq:dynamic_schedule} and \eqref{eq:dw}.
                    \State Compute the perturbation $\boldsymbol{\epsilon}_{\mathrm{SAM}}$ following Eq.~\eqref{eq:sam_perturbation}.
                    \State Set $\boldsymbol{\epsilon}_{\mathrm{NCSAM}}$ using Eq.~\eqref{eq:ncsam_perturbation}.
                    \State Update $\mathbf{w}$ using the gradient of $L_{\mathcal{B}}(\mathbf{w}+\boldsymbol{\epsilon}_{\mathrm{NCSAM}})$.
                \EndIf
            \EndFor
        \EndFor
    \end{algorithmic}
\end{algorithm}
\end{small}

\section{Experiments}
\subsection{Controllable Noise Benchmarks}

%语法修改后
% \textbf{Dataset.} We validate the effectiveness of the proposed NCSAM method on three widely used datasets (i.e., CIFAR-10, CIFAR-100 and Tiny-ImageNet) with synthetic noisy labels. For CIFAR-10 and CIFAR-100, we consider two commonly used noisy label settings: symmetric noise and instance-dependent noise. For the Tiny-ImageNet dataset~\cite{le-2015-tiny}, we adopt symmetric noise and asymmetric noise. Symmetric noise (also known as uniform noise)~\cite{li-2020-dividemix} is generated by uniformly flipping a certain percentage of the original labels to all possible classes. Asymmetric noise is generated by flipping labels only between adjacent (semantically similar) classes with a fixed probability. Instance-dependent noise is generated by assigning incorrect labels based on the visual features of each instance, making the noise distribution dependent on the input samples. Following existing methods~\cite{karim-2022-unicon} \cite{sarfraz-2021-noisy}, we set the symmetric noise rate to 0.2 and the asymmetric noise rate to 0.4 on Tiny-ImageNet. Referring to existing studies~\cite{zhao-2022-centrality}\cite{zhu-2021-second}, we adopt noise rates ranging from 0.2 to 0.8 on the CIFAR-10/100 datasets.

\textbf{Dataset.} 
We first evaluate NCSAM on three widely used noisy-label benchmarks, including CIFAR-10, CIFAR-100, and Tiny-ImageNet. For CIFAR-10 and CIFAR-100, we consider both symmetric noise and instance-dependent noise settings. For Tiny-ImageNet~\cite{le-2015-tiny}, we adopt symmetric noise and asymmetric noise protocols. Symmetric noise(also known as uniform noise)~\cite{li-2020-dividemix} is generated by uniformly flipping labels at random, whereas asymmetric noise only flips labels between semantically similar classes, resulting in more structured class-dependent corruption. Instance-dependent noise dynamically generates corrupted labels conditioned on image features, making the noise distribution dependent on the input samples themselves and therefore substantially more challenging than random corruption.
Following prior work~\cite{karim-2022-unicon} \cite{sarfraz-2021-noisy}, the symmetric noise ratio on Tiny-ImageNet is set to 20\%, while the asymmetric noise ratio is set to 45\%.
For CIFAR-10 and CIFAR-100, we evaluate noise rates ranging from 20\% to 80\%~\cite{zhao-2022-centrality}\cite{zhu-2021-second}.

% \textbf{Experimental setup.} We adopt ResNet-18 as the backbone network and train the model for 200 epochs on the CIFAR and Tiny-ImageNet datasets respectively. In particular, ResNet-32 was used as the backbone network for the CIFAR dataset containing instance noise. We use staged optimization: first, the model is trained for 50 epochs with Stochastic Gradient Descent (SGD) as the optimizer, followed by training with the NCSAM optimizer. The momentum is set to 0.9, the weight decay coefficient is 0.001, and the batch size is 128. The initial learning rate is set to 0.05. For data augmentation, we use the Cutout method \cite{devries-2017-improved} with parameters configured as size=16 and p=0.5.  
\textbf{Experimental setup.} We adopt ResNet-18 as the backbone network and train the model for 200 epochs on the CIFAR-10, CIFAR-100, and Tiny-ImageNet datasets. In particular, PreResNet-32 is used as the backbone for the CIFAR-10/100 datasets under instance-dependent noise. We employ a staged optimization strategy: the model is first trained for 50 epochs using SGD or SAM, then switched to the NCSAM optimizer for the remaining epochs. The momentum is set to 0.9, the weight decay to 0.001, the batch size to 128, and the initial learning rate to 0.05.  All experiments are implemented on a single GeForce RTX 3080 GPU using the PyTorch 2.6.0 framework.

% \textbf{Comparison with SOTA methods}

\textbf{Comparison with Representative Methods.}
The methods marked with * explicitly rely on sample selection or label correction, while NCSAM only modifies the optimization process, we analyze the results by separating these two categories rather than claiming uniform superiority over all methods.

Table~\ref{tab:idn_comparison} compares different methods under instance-dependent noise on CIFAR-10 and CIFAR-100. On CIFAR-10, NCSAM achieves the best result at the 20\% noise rate, reaching 94.51\%, and clearly improves over SAM at all noise levels. However, under 60\% instance-dependent noise, NCSAM obtains 70.11\%, which is lower than strong sample-selection or label-correction methods such as CC, DivideMix, and CAL. This result is reasonable because severe instance-dependent noise on CIFAR-10 tends to corrupt visually ambiguous samples in a concentrated manner; methods that explicitly identify clean samples or correct labels can directly reduce the effective noise ratio, whereas NCSAM keeps all samples and only compensates for the noise-induced perturbation bias during optimization. Therefore, when the corrupted labels dominate the training signal on a low-class dataset, optimization-level correction alone cannot fully replace explicit label cleaning.

The trend on CIFAR-100 is different. Although NCSAM is slightly below CC at the 20\% and 40\% noise rates, it achieves the best accuracy at the 60\% noise rate, reaching 65.46\%, which is 6.06 percentage points higher than CC and 18.74 percentage points higher than DivideMix. This improvement indicates that NCSAM is particularly effective when instance-dependent noise appears in a fine-grained classification setting. CIFAR-100 contains many more classes and stronger inter-class similarity than CIFAR-10, making clean-sample selection and pseudo-label correction less reliable under heavy corruption. In this case, preserving all training samples while correcting the noisy perturbation direction helps NCSAM avoid the error accumulation caused by incorrect sample filtering or pseudo-labeling.

Table~\ref{nc-cifar} reports the results under symmetric label noise. On CIFAR-10, NCSAM achieves the best accuracy across all noise rates, including 89.00\% at 60\% noise and 79.58\% at 80\% noise. These gains show that the proposed compensation mechanism is effective when label corruption is approximately random and the induced noisy gradients can be treated as a perturbation bias to be suppressed. On CIFAR-100, NCSAM obtains the best results at 20\% and 40\% noise, but RegCE performs better at 60\% and 80\% noise. Even so, NCSAM remains substantially stronger than SAM and BSAM, which suggests that the proposed correction stabilizes sharpness-aware optimization, although extremely severe symmetric noise on a 100-class dataset may still require stronger regularization or explicit robust-loss design.

Table~\ref{tiny-imagenet} further evaluates NCSAM on Tiny-ImageNet. Compared with SAM, NCSAM improves the accuracy from 58.48\% to 64.59\% under 20\% symmetric noise and from 39.30\% to 46.11\% under 45\% asymmetric noise. Although DISC achieves the highest overall accuracy because it uses an explicit noisy-label learning strategy, NCSAM consistently improves over the optimization baselines without sample selection, label correction, or additional teacher models. This confirms that correcting the perturbation direction is a useful and lightweight way to improve robustness across both random and structured label noise.

\begin{figure}[t!]
    \centering
    \includegraphics[width=0.95\linewidth]{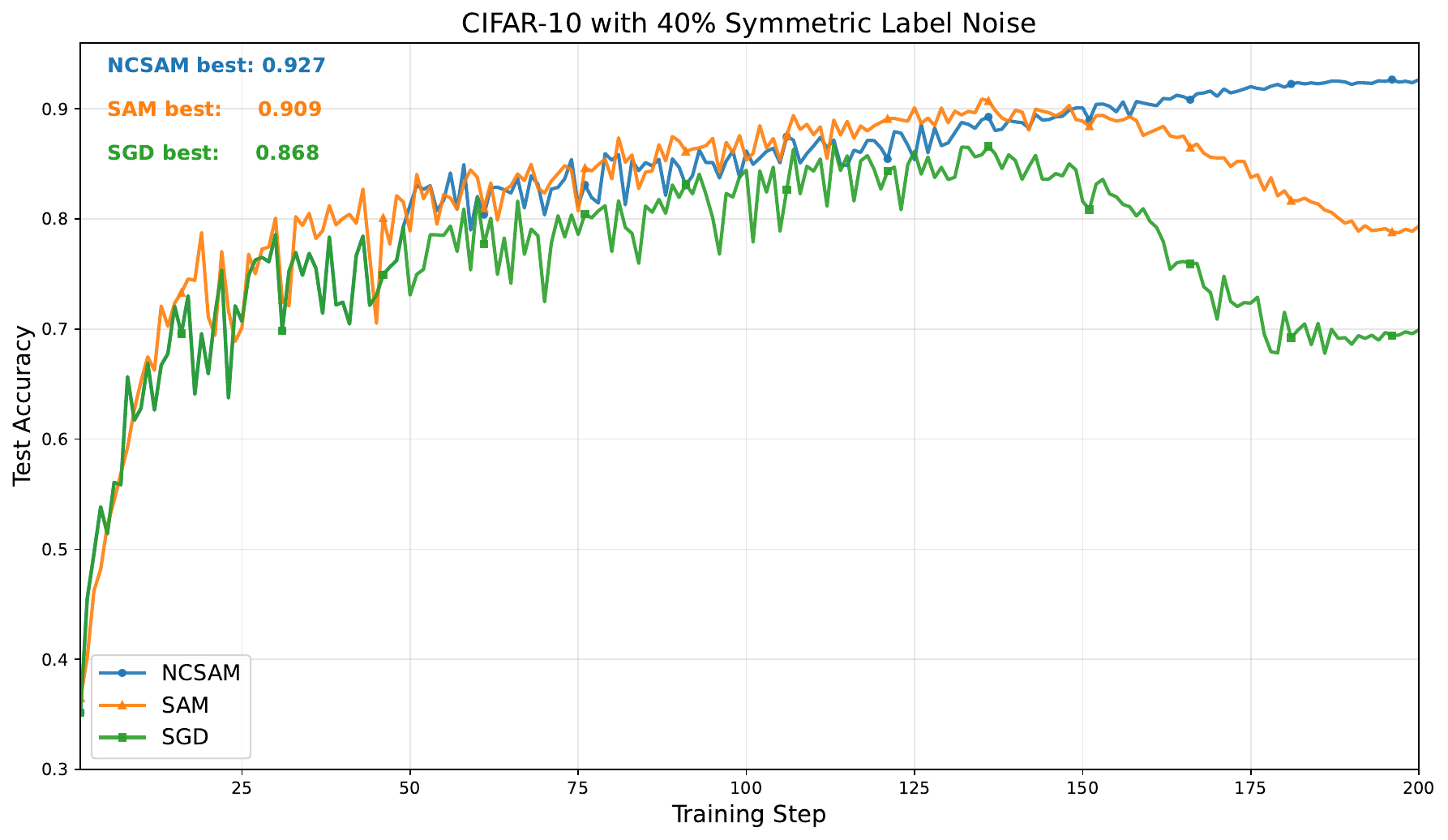}
    \caption{The accuracy curves of ResNet-18 trained on CIFAR-10 with 40\% random label noise using three optimization methods (NCSAM, SAM, and SGD) are shown as a function of training epochs.} 
    \label{fig:Accuracy curve comparison}
\end{figure}

These empirical observations are consistent with the theoretical analysis in Section~\ref{sec:theory}. Label noise distorts the perturbation direction of SAM through mixed gradient terms, causing the optimizer to search for flat regions of the noisy loss landscape rather than flat regions associated with clean semantic information. NCSAM reduces this deviation by introducing an explicit noise-compensation term, thereby improving the stability of sharpness-aware optimization. Figure~\ref{fig:Accuracy curve comparison} further supports this interpretation: while SGD and SAM show later-stage degradation caused by noisy-label memorization, NCSAM maintains a more stable accuracy curve throughout training.

\begin{table*}[t]
\centering
\caption{Comparison with representative methods on CIFAR-10 and CIFAR-100 with instance-dependent noise.
 All the other results are directly from~{\protect\cite{zhao-2022-centrality}}. The results with * use sample selection or label correction for label noise mitigation. Bold numbers indicate the best result within each column.}
\label{tab:idn_comparison}
%\resizebox{\textwidth}{!}{
\begin{tabular}{l|ccc|ccc}
\hline
\multirow{2}{*}{Method} 
& \multicolumn{3}{c|}{CIFAR-10} 
& \multicolumn{3}{c}{CIFAR-100} \\
& Inst. 20\% & Inst. 40\% & Inst. 60\% 
& Inst. 20\% & Inst. 40\% & Inst. 60\% \\
\hline
Forward T* \cite{patrini-2017-making}     & 87.22  & 79.37  & 66.56  & 58.19  & 42.80 & 27.91  \\
Co-teaching* \cite{han-2018-co}   & 88.87  & 73.00  & 62.51 & 43.30 & 23.21  & 12.58 \\
Co-teaching+* \cite{yu-2019-does}  & 89.80  & 73.78 & 59.22  & 41.71  & 24.45  & 12.58 \\
JoCoR* \cite{wei-2020-combating}         & 88.78  & 71.64  & 63.46  & 43.66 & 23.95 & 13.16  \\
Reweight-R* \cite{xia-2019-anchor}    & 90.04  & 84.11  & 72.18  & 58.00 & 43.83  & 36.07  \\
DivideMix* \cite{li-2020-dividemix}     & 93.33 & \textbf{95.07 } & 85.50 & 79.04  & 76.08  & 46.72  \\
CORSES$^2$* \cite{cheng-2020-learning}     & 91.14  & 83.67  & 77.68 & 66.47  & 58.99  & 38.55  \\
CAL* \cite{zhu-2021-second}           & 92.01  & 84.96  & 79.82 & 69.11  & 63.17  & 43.58 \\
CC* \cite{zhao-2022-centrality}            & 93.68  & 94.97 & \textbf{94.95 } & \textbf{79.61 } & \textbf{76.58 } & 59.40  \\
\hline
DMI \cite{patrini-2017-making}           & 88.57  & 82.82  & 69.94  & 57.90 & 42.70  & 26.96  \\
Mixup \cite{zhang-2017-mixup}        & 87.71  & 82.65  & 58.59 & 46.31  & 45.14  & 23.77  \\
GCE \cite{zhang-2018-generalized}          & 89.80 & 78.95& 60.76  & 58.01  & 45.69  & 35.08 \\
Peer Loss \cite{liu-2020-peer}     & 89.12  & 83.26  & 74.53  & 61.16 & 47.23    & 31.71  \\
CE                     & 83.93  & 67.64  & 43.83  & 57.35  & 43.17  & 24.42  \\
SAM 
& 89.94  & 74.72  & 51.65 
& 69.98  & 56.38 & 38.09  \\
\textbf{NCSAM (ours)} 
& \textbf{94.51} & 90.23  & 70.73
& 74.49  & 72.82 & \textbf{64.83} \\
\hline
\end{tabular}
%}
\end{table*}

\begin{table}[h!]
\centering
\footnotesize
\caption{Comparison of test accuracy using different methods on CIFAR-10 and CIFAR-100 datasets with varying noise types and levels. 
The baseline results are taken from~{\protect\cite{kang-2023-unleashing}} and~{\protect\cite{deng-2024-bsam}}. RegCE, SAM, BSAM and NCSAM use ResNet18 as the architecture, whereas other methods in the comparison use PreActResNet18. 
The results with * use sample selection or label correction for label noise mitigation. Bold numbers indicate the best result within each column.}
\label{nc-cifar}
% S列：整数.小数 对齐
\begin{tabular}{cc S[table-format=2.2] S[table-format=2.2] S[table-format=2.2] S[table-format=2.2]}
\toprule
\multicolumn{2}{c}{} & \multicolumn{4}{c}{\textbf{Symmetric Noise Rate}} \\
\textbf{Dataset} & \textbf{Method} & \textbf{20\%} & \textbf{40\%} & \textbf{60\%} & \textbf{80\%} \\
\midrule
\multirow{14}{*}{CIFAR-10}
& Forward*     & 88.87 & 83.28 & 75.15 & 58.58 \\
& Co-teaching* & 92.05 & 87.73 & 85.10 & 44.16 \\
& LIMIT*       & 89.63 & 85.40 & 78.05 & 58.71 \\
& SLN*         & 88.77 & 87.03 & 80.57 & 63.99 \\
& SL*          & 92.45 & 89.22 & 84.63 & 72.59 \\
& APL*         & 92.51 & 89.34 & 85.01 & 70.52 \\
& CTRR*        & 93.05 & 92.16 & 87.34 & 83.66 \\
& RegCE*      & 93.77 & 92.23 & 87.51 & 77.40 \\
\cmidrule(lr){2-6} 
& CE          & 88.51 & 82.73 & 76.26 & 59.25 \\
%& Forward*     & 88.87 & 83.28 & 75.15 & 58.58 \\
& GCE         & 91.22 & 89.26 & 85.76 & 70.57 \\
%& Co-teaching* & 92.05 & 87.73 & 85.10 & 44.16 \\
% & LIMIT*       & 89.63 & 85.40 & 78.05 & 58.71 \\
% & SLN*         & 88.77 & 87.03 & 80.57 & 63.99 \\
% & SL*          & 92.45 & 89.22 & 84.63 & 72.59 \\
% & APL*         & 92.51 & 89.34 & 85.01 & 70.52 \\
% & CTRR*        & 93.05 & 92.16 & 87.34 & 83.66 \\
% & RegCE*      & 93.77 & 92.23 & 87.51 & 77.40 \\
& SGD         & 85.94 & 70.25 & 48.50 & 28.91 \\
& SAM         & 90.44 & 79.71 & 67.22 & 76.94 \\
& BSAM        & 93.34 & 85.58 & 76.37 & 78.27 \\
& \textbf{NCSAM}       & \textbf{94.53} & \textbf{92.65} & \textbf{89.00} & \textbf{79.58} \\
\midrule
\multirow{14}{*}{CIFAR-100}
& Forward*     & 58.72 & 50.10 & 39.35 & 17.15 \\
& Co-teaching* & 57.46 & 57.64 & 31.59 & 15.28 \\
& LIMIT*       & 58.02 & 49.71 & 37.05 & 20.01 \\
& SLN*         & 55.35 & 51.39 & 35.53 & 11.96 \\
& SL*         & 66.46 & 61.44 & 54.17 & 34.22 \\
& APL*         & 68.09 & 63.46 & 53.63 & 20.00 \\
& CTRR*        & 70.09 & 65.32 & 54.20 & 43.69 \\
& RegCE*       & 74.84 & 70.18 & \textbf{62.91} & \textbf{47.07} \\
\cmidrule(lr){2-6} 
& CE          & 60.57 & 52.48 & 43.20 & 22.96 \\
%& Forward*     & 58.72 & 50.10 & 39.35 & 17.15 \\
& GCE         & 68.31 & 62.25 & 53.86 & 19.31 \\
% & Co-teaching* & 57.46 & 57.64 & 31.59 & 15.28 \\
% & LIMIT*       & 58.02 & 49.71 & 37.05 & 20.01 \\
% & SLN*         & 55.35 & 51.39 & 35.53 & 11.96 \\
% & SL*         & 66.46 & 61.44 & 54.17 & 34.22 \\
% & APL*         & 68.09 & 63.46 & 53.63 & 20.00 \\
% & CTRR*        & 70.09 & 65.32 & 54.20 & 43.69 \\
% & RegCE*       & 74.84 & 70.18 & \textbf{62.91} & \textbf{47.07} \\
& SGD         & 65.42 & 48.91 & 31.46 & 12.32 \\
& SAM         & 69.10 & 55.06 & 35.59 &  9.82 \\
& BSAM        & 71.65 & 57.61 & 38.71 & 32.69 \\
& \textbf{NCSAM}       & \textbf{76.05} & \textbf{70.59} & 58.01 & 39.02\\
\bottomrule
\end{tabular}
\end{table}

% \begin{table}[h!]
% \centering
% \caption{Comparison with the different methods on Tiny ImageNet
% with sym. and asym. The baseline results are taken from~{\protect\cite{sarfraz-2021-noisy}}. SAM, SGD and NCSAM use ResNet18 as the architecture, whereas all other methods in the comparison use PreActResNet18. 
% The best results are highlighted in bold.}
% \resizebox{0.5\textwidth}{!}{
% \begin{tabular}{cccccc}
% \toprule
% %\multicolumn{2}{c}{} \\%& \multicolumn{4}{c}{\textbf{sym rate}} \\
% \textbf{Dataset} & \textbf{Method} & \textbf{sym rate 20\%} &\textbf{asym rate 40\%} \\
% \midrule
% \multirow{3}{*}{Tiny-imagenet}
% & Standard  & 35.8  & -\\
% & Decoupling  & 37.0  & -\\
% & Co-teaching+  & 48.2  & -\\
% & M-Correction  & 57.2  & -\\
% & NCT  & 58.2  & -\\
% & UNICON  & 59.2  & -\\
% & SGD  & 53.57  & 43.04\\
% & SAM  & 58.48 & 44.46 \\
% & NCSAM& \textbf{64.59} &\textbf{51.87}\\
% \bottomrule\label{tiny-imagenet}
% \end{tabular}
% }
% \end{table}

\begin{table}[h!]
\centering
\footnotesize % 缩小字体，适配论文排版
\caption{Comparison with different methods on Tiny ImageNet under symmetric and asymmetric label noise. Baseline results are taken from \protect\cite{sarfraz-2021-noisy}\protect\cite{li-2023-disc}. SAM and NCSAM adopt ResNet18 as the backbone, while all other compared methods use PreActResNet18. The results with * use sample selection or label correction for label noise mitigation. Bold numbers indicate the best result within each comparison group.}
\label{tiny-imagenet}
\begin{tabular}{cc S[table-format=2.2] S[table-format=2.2]}
\toprule
\textbf{Dataset} & \textbf{Method} & \textbf{Sym 20\%} & \textbf{Asym 45\%} \\
\midrule
\multirow{9}{*}{Tiny-ImageNet}
& Decoupling*   & 37.00 & {26.6} \\
& F-correction* & 44.5 & {0.67} \\
& MentorNet*    & 45.7 & {26.61} \\
& Co-teaching+* & 48.20 & {26.87} \\
& M-Correction* & 57.20 & {24.8} \\
& UNICON*       & 59.20 & {--} \\
& DISC*        & \textbf{67.9} & \textbf{53.6} \\
\cmidrule(lr){2-4} 
& Standard     & 35.80 & {26.3} \\
& NCT          & 58.20 & {43} \\
& SAM          & 58.48 & 39.30 \\
& \textbf{NCSAM}        & {64.59} & {46.11} \\
\bottomrule
\end{tabular}
\end{table}

\subsection{Real-world Noise Benchmarks}
% \textbf{Dataset.} Food-101N~\cite{bossard14} is a benchmark containing 101 food categories. It consists of 75,000 noisy-label training images and 25,000 manually annotated testing images. Animal-10N~\cite{song-2019-selfie} is a web-crawled benchmark with 5 pairs of confusing animals, containing 50,000 training images and 5,000 testing images. Clothing1M~\cite{xiao-2015-learning} is a large-scale crawled noisy clothing dataset with 1 million training images and 10,000 testing images obtained from several online shopping websites. It has 14 classes with an estimated noise rate of approximately 38.5\%.
\textbf{Dataset.} To further evaluate the generalization ability of NCSAM in realistic noisy environments, we conduct experiments on two real-world noisy-label datasets: Animals-10N~\cite{song-2019-selfie} and Food-101N~\cite{bossard14}.
Unlike synthetic noisy labels, the corrupted annotations in these datasets originate from web crawling or human annotation errors.
As a result, their noise distributions are substantially more complicated and usually contain mixed corruption patterns, including annotation ambiguity, semantic overlap, open-set noise, and distribution mismatch.
Therefore, real-world noisy-label learning is generally more challenging than controllable synthetic corruption.

\textbf{Experimental setup.} Following~\cite{chen-2019-understanding}\cite{song-2019-selfie}, for Animal-10N we use VGG-19~\cite{simonyan-2014-very} (not pretrained on ImageNet) as the backbone. Following~\cite{li-2020-dividemix}\cite{zhang-2021-learning}, for Food-101N we use ResNet-50~\cite{he-2016-deep} (pretrained on ImageNet) as the backbone. The training epochs are 120 for Animal-10N and 100 for the others. We use the SAM optimizer with an initial learning rate of 0.05 for all datasets: weight decay = $5 \times 10^{-4}$ for Animal-10N and Food-101N. The batch size is set to 64 for Animal-10N and 32 for Food-101N.

\textbf{Comparison with Representative Methods.} Tables \ref{animals-10n} and \ref{food-101} present the experimental results on Animals-10N and Food-101N, respectively.
As observed, NCSAM consistently outperforms standard SAM across all real-world noisy datasets.
Compared with the synthetic noise setting, the performance gains of NCSAM on real-world noise are relatively smaller. We attribute this phenomenon to the fact that real-world label noise often consists of heterogeneous and mixed noise patterns, where erroneous labels do not follow a unified statistical distribution. Consequently, the induced noisy gradients exhibit more complex characteristics and are harder to estimate accurately.
Nevertheless, NCSAM still achieves consistent improvements without relying on sample selection, label correction, or additional teacher networks. This suggests that even when the underlying noise structure cannot be explicitly characterized, mitigating the dominant noise-induced bias in the perturbation direction remains effective for improving the stability and generalization capability of sharpness-aware optimization under noisy-label learning.

% \begin{table}[h!]
% \centering
% \caption{Performance comparison on real-world noisy datasets.}
% \begin{tabular}{cccc}
% \toprule
% %\multicolumn{2}{c}{} \\
% \textbf{Dataset} &\textbf{model} &\textbf{Method} & \textbf{Accuracy}  \\
% \midrule
% \multirow{2}{*}{\centering Animal-10N} & \multirow{2}{*}{\centering VGG-19}
% & SAM  & 81.96 \\
% && NCSAM& \textbf{84.32} \\  %f0.1 gn系数上限为0.05
% \midrule
% \multirow{2}{*}{\centering Food101N} & \multirow{2}{*}{\centering ResNet-50}
% & SAM  & 86.54 \\
% && NCSAM& \textbf{86.93} \\ %f0.2gn系数上限为0.1
% \midrule
% \multirow{2}{*}{\centering Clothing1M} & \multirow{2}{*}{\centering ResNet-50}
% & SAM  & 70.19 \\
% && NCSAM& \textbf{70.73} \\ %f0.3 gn系数0.2
% \bottomrule\label{Real-world Noise}
% \end{tabular}
% \end{table}

\begin{table}[h!]
\centering
\footnotesize % 缩小字体，适配论文排版
\caption{Comparison with different methods on Animals-10N. The results with * use sample selection or label correction for label noise mitigation. Other results are taken directly from \protect\cite{li-2023-disc}. Bold numbers indicate the best result within each comparison group.}
\label{animals-10n}
\begin{tabular}{cc S[table-format=2.2]}
\toprule
\textbf{Dataset} & \textbf{Method} & \textbf{Accuracy (\%)} \\
\midrule
\multirow{9}{*}{Animals-10N}
& SELFIE*     & 81.80 \\
& PLC*         & 83.40 \\
& Nested Co-teaching*  & 84.10 \\
& DISC*       & \textbf{87.10} \\
\cmidrule(lr){2-3} 
& CE          & 79.40 \\
& GCE       & 81.50 \\
%& SELFIE*     & 81.80 \\
& Mixup     & 82.70 \\
& Co-learning & 83.00 \\
%& PLC*         & 83.40 \\

& GJS        & 84.20 \\
%& DISC*       & \textbf{87.10} \\
& SAM       & 81.96 \\
& \textbf{NCSAM}       & 84.32 \\
\bottomrule
\end{tabular}
\end{table}

% 表格2：Food-101 数据集
\begin{table}[h!]
\centering
\footnotesize % 缩小字体，适配论文排版
\caption{Comparison with different methods on Food-101. The results with * use sample selection or label correction for label noise mitigation. Other results are taken directly from \protect\cite{li-2023-disc}. Bold numbers indicate the best result within each comparison group.}
\label{food-101}
\begin{tabular}{cc S[table-format=2.2]}
\toprule
\textbf{Dataset} & \textbf{Method} & \textbf{Accuracy (\%)} \\
\midrule
\multirow{8}{*}{Food-101N}
& CleanNet*   & 83.95 \\
& PLC*        & 83.40 \\
& DISC*      & \textbf{89.02} \\
\cmidrule(lr){2-3} 
& CE          & 81.67 \\
& GCE       & 85.83 \\
& GJS        & 86.56 \\
& Mixup     & 87.34 \\
& Co-learning & 87.57 \\
& SAM       & 86.54 \\
& \textbf{NCSAM}       & 86.93 \\
\bottomrule
\end{tabular}
\end{table}

\subsection{Ablation Study}
% To investigate the contribution of each component of the proposed NCSAM framework, we conduct a series of ablation experiments on CIFAR-10 and CIFAR-100 under synthetic noisy labels. Specifically, we analyze the effectiveness of three key modules:
To further analyze the contribution of different components in NCSAM, we conduct extensive ablation studies on CIFAR-10 and CIFAR-100.
Specifically, we investigate:(1) the effect of progressive label-flipping ratios,
(2) the influence of dynamic noise-compensation strategies,
and (3) the impact of the maximum compensation strength.
\subsubsection{Effect of Progressive Label-Flipping Ratio}
The progressive label-flipping mechanism is designed to approximate the dominant direction of noisy gradients.This experiment investigates how different label-flip ratios influence the effectiveness of perturbation compensation.
We conduct experiments on CIFAR-100 under 40\% and 60\% symmetric noisy labels. The label-flip ratio $r_{\mathrm{flip}}$ varies from 20\% to 70\%.
Samples are selected based on the proposed logit-gap criterion, while the flipped target label is assigned to the second-largest logit class.All remaining settings are kept identical to the main experiments.

% \paragraph{Results.} As shown in Table~\ref{tab:label-flip}, the optimal progressive label-flipping ratio depends on the underlying noise level. Under moderate noise (40\% symmetric noise), the best performance is achieved when the flip ratio matches the noise rate ($  r_{\rm flip}=40\%  $). In contrast, under higher noise (60\% symmetric noise), a relatively smaller flip ratio ($  r_{\rm flip}=40\%  $) yields better performance than using a flip ratio equal to the noise rate. These results indicate that when the noise level is low to moderate, matching the flip ratio to the noise rate leads to better alignment between the simulated noise and the underlying label corruption, whereas under severe noise, a lower flip ratio avoids excessive disturbance and improves training stability.

As shown in Table~\ref{tab:label-flip}, the optimal label-flip ratio strongly depends on the underlying noise level.Under moderate noise conditions (40\%), the best performance is achieved when the simulated flip ratio approximately matches the actual noise rate.This indicates that the generated noisy gradients can effectively approximate the perturbation bias induced by real noisy supervision.
However, under severe noise conditions (60\%), smaller flip ratios surprisingly produce better performance.We conjecture that when the original supervision is already heavily corrupted, excessively strong simulated perturbations may further increase optimization instability.Therefore, effective noise compensation should maintain a balance between perturbation correction and optimization stability rather than simply increasing the compensation strength.

\begin{table}[h!]
\centering
\caption{Effect of different progressive label-flip ratios on CIFAR-100 under Symmetric noise noise.}
\label{tab:label-flip}
\resizebox{0.5\textwidth}{!}{
\begin{tabular}{l|cccccc}
\hline
\multirow{2}{*}{Symmetric noise} 
& \multicolumn{6}{c}{Flip rate} \\
 & 20\% & 30\% & 40\% & 50\% & 60\% & 70\%  \\
\hline
Cifar100(40\%) & 69.03  & 69.41 & \textbf{70.65}  & 68.93 &68.11 & - \\
Cifar100(60\%) & -   & 61.78  &\textbf{61.80} &60.99 &59.70 &57.29  \\
\hline
\end{tabular}
}
\end{table}

\subsubsection{Effect of Dynamic Noise Compensation Strategies}
\label{sec:coeff_analysis}

The dynamic coefficient $s(t)$ controls the strength of the proposed noise-compensation term throughout training.Unlike conventional warm-up strategies that only adjust the learning rate, $s(t)$ directly determines the influence of perturbation compensation at different training stages.

\paragraph{Effect of Different Scheduling Strategies.}
% We first examine the influence of different scheduling strategies for $s(t)$. Specifically, we compare:
% (1) \emph{constant scaling}, where $s(t)=c$ for all training epochs;
% (2) \emph{progressive warm-up}, where $s(t)$ increases smoothly from zero and saturates after a predefined stage.
We compare two compensation strategies:
(1) constant scaling and
(2) progressive linear warm-up.
As shown in Table~\ref{tiny}, constant compensation generally leads to inferior performance and unstable optimization across multiple datasets.
This is because during the early stage of training, the model has not yet learned sufficiently discriminative semantic representations.
Consequently, the simulated noisy gradients are still unreliable.
Aggressive compensation applied too early may incorrectly push the optimization trajectory away from the true clean-gradient direction.
In contrast, progressive warm-up gradually increases the compensation strength as the model becomes more semantically stable, thereby producing substantially more robust optimization behavior.
These results suggest that perturbation compensation should only become dominant after the model acquires sufficiently reliable semantic representations.

\begin{table}[h!]
\centering
\caption{Effect of different scaling strategies for noise compensation in NCSAM.}
\begin{tabular}{cccccc}
\toprule
%\multicolumn{1}{c}{} \\%& \multicolumn{4}{c}{\textbf{sym rate}} \\
\textbf{Variants} & \textbf{cifar10n(20\%)} & \textbf{cifar100n(20\%)}  \\
\midrule
%\multirow{3}{*}{Tiny-imagenet}
Constant & 94.41  & 53.57  \\
Linear warm-up& \textbf{94.55}  & \textbf{76.05} \\
\bottomrule\label{tiny}
\end{tabular}
\end{table}

%  In Table \ref{tiny},the experimental results on CIFAR-10 and CIFAR-100 with symmetric noise show that constant scaling leads to unstable optimization and inferior performance,especially when the coefficient is large. This is because the simulated noise gradient $g^*_{noise}$ is unreliable in the early stage of training, and aggressive compensation may push the optimization trajectory away from the clean-gradient direction.

% In contrast, progressive schedules consistently yield higher accuracy and improved stability, confirming that noise-gradient compensation should only be activated after the model has learned a reasonably discriminative representation.

% \paragraph{Impact of the Maximum Scaling Magnitude.}
% We further investigate the effect of the upper bound $\kappa$ of $s(t)$, which determines the maximum strength of noise compensation during the later stage of training. As shown in Table~\ref{tab:upper bound}, the optimal value of $\kappa$ varies significantly across datasets.

\subsubsection{Effect of Maximum Compensation Strength}

We further investigate the influence of the upper bound $\kappa$ of the dynamic compensation coefficient.
As shown in Table~\ref{tab:upper bound}, the optimal compensation strength varies significantly across different datasets and noise structures.

Under symmetric noise, label corruption is approximately random, causing the noisy gradients to behave largely stochastically.
In this case, relatively small compensation strengths are usually sufficient, whereas excessively large coefficients may instead introduce additional optimization bias.

Under asymmetric noise and instance-dependent noise, corrupted labels exhibit stronger structural consistency and directional perturbation bias.
Therefore, larger compensation strengths become more effective at correcting systematic gradient deviation.

For real-world noisy datasets, the underlying corruption patterns are significantly more complicated and difficult to model accurately.
Overly aggressive compensation may further amplify estimation errors, making conservative upper bounds empirically more stable.

Overall, the choice of the compensation strength should jointly consider:
(1) the reliability of noisy-gradient estimation,
(2) the structural consistency of the noise,
and (3) the stability of the optimization process.

As the noise level increases, moderately increasing the compensation strength can be beneficial, provided that the optimization process remains stable.

%还需要添加做实验 不同的噪声比如现在只做了对称噪声，还有非对称噪声和实例噪声也需要做实验表明其gn系数的上限设置

\begin{table}[h!]
\centering
\footnotesize
\caption{The average accuracy of the last five rounds demonstrates the impact of dynamic noise-compensation strength on model robustness.}
\label{tab:upper bound}
\resizebox{0.5\textwidth}{!}{
\begin{tabular}{l|cccccc}
\hline
\multirow{2}{*}{\textbf{Dataset}} 
& \multicolumn{6}{c}{\textbf{upper bound of $s(t)$}} \\
 & 0.03 & 0.05 & 0.1 &0.2 & 0.3 & 0.4  \\
\hline
Cifar10(20\%sym) & -  & 94.49 & \textbf{94.55}  & 93.91 &- & -  \\
Cifar100(20\%sym) &   -  &  -  &-  &74.40 &\textbf{75.94} &75.48  \\
Animal10N & 83.58  &\textbf{84.11}   &82.68 &-  &-  & -  \\
Food101N &  -   &  86.82 &\textbf{86.91} &84.53 & - & -  \\
\hline
\end{tabular}
}
\end{table}

\section{Conclusion}
This work systematically investigates the impact of noisy labels on sharpness-aware optimization and shows that label noise distorts both the direction and the magnitude of SAM perturbations, thereby weakening SAM's ability to locate flat minima. Through PAC-Bayesian analysis, we further show that the parameter deviation induced by label noise has a compensable relationship with the perturbation term, which provides a principled foundation for designing noise-robust perturbation-based optimizers.

Motivated by this analysis, we propose NCSAM, a novel optimization framework that explicitly models and compensates for the gradient bias introduced by noisy labels. NCSAM generates structured noise gradients via progressive noise simulation and applies a dynamically scaled correction term to realign perturbed updates with clean-gradient directions while stabilizing the effective perturbation magnitude. Without relying on sample selection or label correction, NCSAM suppresses the memorization of noisy labels and preserves the flatness-promoting behavior of SAM.

Extensive experiments on synthetic and real-world noisy datasets show that NCSAM consistently improves over SAM-style optimization baselines and remains competitive with representative noisy-label learning methods. More broadly, this work suggests a new perspective on noise-robust optimization: correcting perturbation distortion, rather than relying solely on elaborate label-cleaning pipelines, can be both theoretically principled and empirically effective. We hope this insight inspires future research on perturbation-based training and robust optimization under weak supervision.

% 附录开始
\appendix % 关键命令：自动将后续章节编号改为A、B、C...
\renewcommand{\theequation}{\Alph{section}.\arabic{equation}} % 附录公式独立编号：A.1, B.1...
\renewcommand{\thefigure}{\Alph{section}.\arabic{figure}} % 附录图独立编号
\renewcommand{\thetable}{\Alph{section}.\arabic{table}} % 附录表独立编号

\section{Proofs}
\label{app:proofs}
\setcounter{equation}{0}
\renewcommand{\theequation}{A.\arabic{equation}}

\subsection{Complete Derivation of Theorem~\ref{thm:noisy_sam_bound}}
\label{app:noisy_sam_bound}
This appendix gives the complete proof of Theorem~\ref{thm:noisy_sam_bound}. The derivation follows Appendix A.1 of SAM~\cite{foret-2020-sharpness}; the only additional step is the algebraic conversion from the clean empirical risk to the mixed empirical risk. Throughout the proof, the clean/noisy decomposition is used only for analysis. Conditioning on the noisy subset $\hat{\mathcal{S}}$, the PAC-Bayesian probability statement is taken over the clean sample $\mathcal{S}$ drawn from the target distribution $\mathcal{D}$. This conditional statement is not intended to assume that the algorithm knows $\hat{\mathcal{S}}$; it only fixes the corrupted portion so that the empirical term in the PAC-Bayesian bound can be evaluated on samples from the clean target distribution. As stated in Corollary~\ref{cor:unconditional_bound}, the same bound admits an unconditional interpretation whenever the conditional guarantee holds uniformly for each realized noisy subset with fixed $n_s$ and $n_n$.

Let $\mathcal{S}$ be the clean subset with $n_s$ samples, $\hat{\mathcal{S}}$ be the noisy subset with $n_n$ samples, and $\mathcal{M}=\mathcal{S}\cup\hat{\mathcal{S}}$ with $n=n_s+n_n$. The mixed empirical risk is
\begin{equation}
L_{\mathcal{M}}(\mathbf{w})
=
\frac{1}{n}
\left(
n_sL_{\mathcal{S}}(\mathbf{w})+n_nL_{\hat{\mathcal{S}}}(\mathbf{w})
\right).
\label{eq:app_mixed_risk}
\end{equation}
Since the target distribution is the clean distribution $\mathcal{D}$, we start from the PAC-Bayesian bound on $\mathcal{S}$. Following the McAllester bound used in SAM Appendix A.1~\cite{mcallester-1998-some,dziugaite-2017-computing}, for any prior $P$ independent of $\mathcal{S}$ and any posterior $Q$, with probability at least $1-\delta$,
\begin{equation}
\mathbb{E}_{\mathbf{w}'\sim Q}[L_{\mathcal{D}}(\mathbf{w}')]
\le
\mathbb{E}_{\mathbf{w}'\sim Q}[L_{\mathcal{S}}(\mathbf{w}')]
+
\sqrt{
\frac{
D_{\mathrm{KL}}(Q\|P)+\log\frac{n_s}{\delta}
}{
2(n_s-1)
}
}.
\label{eq:app_pac_bayes}
\end{equation}
Here $D_{\mathrm{KL}}(Q\|P)$ denotes the Kullback--Leibler divergence from $Q$ to $P$.

Set $Q=\mathcal{N}(\mathbf{w},\sigma^2\mathbf{I}_K)$ and let $\boldsymbol{\epsilon}\sim\mathcal{N}(\mathbf{0},\sigma^2\mathbf{I}_K)$. Then $\mathbf{w}+\boldsymbol{\epsilon}\sim Q$, and Eq.~\eqref{eq:app_pac_bayes} becomes
\begin{equation}
\mathbb{E}_{\boldsymbol{\epsilon}}[L_{\mathcal{D}}(\mathbf{w}+\boldsymbol{\epsilon})]
\le
\mathbb{E}_{\boldsymbol{\epsilon}}[L_{\mathcal{S}}(\mathbf{w}+\boldsymbol{\epsilon})]
+
\sqrt{
\frac{
D_{\mathrm{KL}}(Q\|P)+\log\frac{n_s}{\delta}
}{
2(n_s-1)
}
}.
\label{eq:app_gaussian_pac_bayes}
\end{equation}

From Eq.~\eqref{eq:app_mixed_risk}, for every perturbation $\boldsymbol{\epsilon}$,
\begin{equation}
L_{\mathcal{S}}(\mathbf{w}+\boldsymbol{\epsilon})
=
\frac{n}{n_s}L_{\mathcal{M}}(\mathbf{w}+\boldsymbol{\epsilon})
-
\frac{n_n}{n_s}L_{\hat{\mathcal{S}}}(\mathbf{w}+\boldsymbol{\epsilon}).
\label{eq:app_clean_exact}
\end{equation}
Because the loss is non-negative, $L_{\hat{\mathcal{S}}}(\mathbf{w}+\boldsymbol{\epsilon})\ge0$, and therefore
\begin{equation}
L_{\mathcal{S}}(\mathbf{w}+\boldsymbol{\epsilon})
\le
\frac{n}{n_s}L_{\mathcal{M}}(\mathbf{w}+\boldsymbol{\epsilon})
=
\frac{1}{1-\alpha}L_{\mathcal{M}}(\mathbf{w}+\boldsymbol{\epsilon}).
\label{eq:app_clean_upper}
\end{equation}
Taking expectation over $\boldsymbol{\epsilon}$ and substituting into Eq.~\eqref{eq:app_gaussian_pac_bayes} yields
\begin{equation}
\mathbb{E}_{\boldsymbol{\epsilon}}[L_{\mathcal{D}}(\mathbf{w}+\boldsymbol{\epsilon})]
\le
\frac{n}{n_s}\mathbb{E}_{\boldsymbol{\epsilon}}[L_{\mathcal{M}}(\mathbf{w}+\boldsymbol{\epsilon})]
+
\sqrt{
\frac{
D_{\mathrm{KL}}(Q\|P)+\log\frac{n_s}{\delta}
}{
2(n_s-1)
}
}.
\label{eq:app_mixed_expected}
\end{equation}

We next use the same KL treatment as SAM Appendix A.1. Let
$P=\mathcal{N}(\boldsymbol{\mu}_P,\sigma_P^2\mathbf{I}_K)$ and $Q=\mathcal{N}(\boldsymbol{\mu}_Q,\sigma_Q^2\mathbf{I}_K)$. Then
\begin{equation}
D_{\mathrm{KL}}(Q\|P)
=
\frac12
\left[
\frac{K\sigma_Q^2+\|\boldsymbol{\mu}_P-\boldsymbol{\mu}_Q\|_2^2}{\sigma_P^2}
-K
+K\log\left(\frac{\sigma_P^2}{\sigma_Q^2}\right)
\right].
\label{eq:app_gaussian_kl}
\end{equation}
Following SAM Appendix A.1, set $\boldsymbol{\mu}_P=\mathbf{0}$, $\boldsymbol{\mu}_Q=\mathbf{w}$, and $\sigma_Q=\sigma$. Since the prior variance must be chosen independently of the clean sample, we consider a discrete grid of admissible prior variances before observing $\mathcal{S}$. Applying a union bound over this grid preserves the PAC-Bayesian validity for all grid values simultaneously. Selecting the best grid element after the bound is established gives
\begin{equation}
D_{\mathrm{KL}}(Q\|P)
\le
\frac12
\left[
1+
K\log\left(1+\frac{\|\mathbf{w}\|_2^2}{K\sigma^2}\right)
\right],
\label{eq:app_kl_bound}
\end{equation}
and the corresponding confidence penalty is upper bounded by
\begin{equation}
\log\frac{n_s}{\delta_j}
\le
\log\frac{n_s}{\delta}+2\log(6n_s+3K).
\label{eq:app_union_penalty}
\end{equation}
Substituting Eqs.~\eqref{eq:app_kl_bound} and \eqref{eq:app_union_penalty} into Eq.~\eqref{eq:app_mixed_expected} gives
\begin{equation}
\begin{split}
\mathbb{E}_{\boldsymbol{\epsilon}}[L_{\mathcal{D}}(\mathbf{w}+\boldsymbol{\epsilon})]
\le& \frac{n}{n_s}\mathbb{E}_{\boldsymbol{\epsilon}}[L_{\mathcal{M}}(\mathbf{w}+\boldsymbol{\epsilon})] + \sqrt{\frac{\mathcal{X}}{n_s-1}},\\
\mathcal{X}=&\frac14K\log\left(1+\frac{\|\mathbf{w}\|_2^2}{K\sigma^2}\right)+\frac14 \\
&+\log\frac{n_s}{\delta}+2\log(6n_s+3K).
\end{split}
\label{eq:app_after_kl}
\end{equation}

It remains to convert the Gaussian expectation into the SAM neighborhood maximum. By the Laurent-Massart chi-squared tail bound~\cite{laurent-2000-adaptive}, for any $u>0$,
\begin{equation}
\Pr\left(
\|\boldsymbol{\epsilon}\|_2^2-K\sigma^2
\ge
2\sigma^2\sqrt{Ku}+2u\sigma^2
\right)
\le
e^{-u}.
\label{eq:app_laurent_massart}
\end{equation}
Taking $u=\log\sqrt{n_s}$ gives, with probability at least $1-1/\sqrt{n_s}$,
\begin{equation}
\begin{split}
\|\boldsymbol{\epsilon}\|_2^2
\le&\sigma^2\left(K+2\sqrt{K\log\sqrt{n_s}}+2\log\sqrt{n_s}\right) \\
&\le\sigma^2K\left(1+\sqrt{\frac{\log n_s}{K}}\right)^2.
\label{eq:app_radius_bound} 
\end{split}
\end{equation}
Choose
\begin{equation}
\sigma
=
\frac{\rho}{
\sqrt{K}
\left(
1+\sqrt{\frac{\log n_s}{K}}
\right)
}.
\label{eq:app_sigma_rho}
\end{equation}
Then $\|\boldsymbol{\epsilon}\|_2\le\rho$ with probability at least $1-1/\sqrt{n_s}$. Let
$A=\{\|\boldsymbol{\epsilon}\|_2\le\rho\}$. Since $0\le L_{\mathcal{M}}\le1$,
\begin{equation}
\begin{split}
\mathbb{E}_{\boldsymbol{\epsilon}}[L_{\mathcal{M}}(\mathbf{w}+\boldsymbol{\epsilon})]
&=
\Pr(A)\mathbb{E}[L_{\mathcal{M}}(\mathbf{w}+\boldsymbol{\epsilon})\mid A] \\
&+\Pr(A^c)\mathbb{E}[L_{\mathcal{M}}(\mathbf{w}+\boldsymbol{\epsilon})\mid A^c]\\
&\le
\max_{\|\boldsymbol{\epsilon}\|_2\le\rho}L_{\mathcal{M}}(\mathbf{w}+\boldsymbol{\epsilon})
+\frac1{\sqrt{n_s}}.
\end{split}
\label{eq:app_expect_to_max}
\end{equation}
Combining Eq.~\eqref{eq:app_expect_to_max} with Eq.~\eqref{eq:app_after_kl} and using
\begin{equation}
\frac{\|\mathbf{w}\|_2^2}{K\sigma^2}
=
\frac{\|\mathbf{w}\|_2^2}{\rho^2}
\left(
1+\sqrt{\frac{\log n_s}{K}}
\right)^2
\label{eq:app_sigma_substitution}
\end{equation}
gives
\begin{equation}
\begin{split}
\mathbb{E}_{\boldsymbol{\epsilon}}[L_{\mathcal{D}}(\mathbf{w}+\boldsymbol{\epsilon})]
\leq& \frac{n}{n_s}\max_{\|\boldsymbol{\epsilon}\|_2\le\rho}\!L_{\mathcal{M}}(\mathbf{w}+\boldsymbol{\epsilon}) + \frac{n}{n_s\sqrt{n_s}} + \sqrt{\frac{\mathcal{X}}{n_s-1}},\\
\mathcal{X}=&\frac14K\log\left(\!1+\frac{\|\mathbf{w}\|_2^2}{\rho^2}\left(1+\sqrt{\frac{\log n_s}{K}}\right)^{\!2}\right)+\frac14 \\
&+\log\frac{n_s}{\delta}+2\log(6n_s+3K).
\end{split}
\label{eq:app_expected_final}
\end{equation}
Finally, SAM Theorem 2 assumes
\begin{equation}
L_{\mathcal{D}}(\mathbf{w})
\le
\mathbb{E}_{\boldsymbol{\epsilon}}[L_{\mathcal{D}}(\mathbf{w}+\boldsymbol{\epsilon})].
\label{eq:app_sam_assumption}
\end{equation}
Using Eq.~\eqref{eq:app_sam_assumption} and noting $n/n_s=1/(1-\alpha)$, we obtain Theorem~\ref{thm:noisy_sam_bound}.

\subsection{Gaussian Tail Bound Used in the Proof}
\label{app:gaussian_tail}
For completeness, we state the concentration result used in Appendix~\ref{app:noisy_sam_bound}. Following Laurent and Massart~\cite{laurent-2000-adaptive}, if $\boldsymbol{\epsilon}\sim\mathcal{N}(\mathbf{0},\sigma^2\mathbf{I}_K)$, then for any $u>0$,
\begin{equation}
\Pr\left(
\|\boldsymbol{\epsilon}\|_2^2-K\sigma^2
\ge
2\sigma^2\sqrt{Ku}+2u\sigma^2
\right)
\le
e^{-u}.
\label{eq:app_tail_statement}
\end{equation}
Equivalently, with probability at least $1-e^{-u}$,
\begin{equation}
\|\boldsymbol{\epsilon}\|_2^2
\le
\sigma^2\left(K+2\sqrt{Ku}+2u\right).
\label{eq:app_tail_equivalent}
\end{equation}
The proof is standard and is not repeated here; this is the same tail bound used in SAM Appendix A.1~\cite{foret-2020-sharpness}.

Taking $u=\log\sqrt{n_s}$ gives $e^{-u}=1/\sqrt{n_s}$. Hence, with probability at least $1-1/\sqrt{n_s}$,
\begin{equation}
\|\boldsymbol{\epsilon}\|_2^2
\le
\sigma^2
\left(
K+2\sqrt{K\log\sqrt{n_s}}+2\log\sqrt{n_s}
\right).
\label{eq:app_tail_ns}
\end{equation}
SAM Appendix A.1 upper bounds the right-hand side by the simpler radius
\begin{equation}
\sigma^2K
\left(1+\sqrt{\frac{\log n_s}{K}}\right)^2.
\label{eq:app_tail_sam_radius}
\end{equation}
Thus, choosing
\begin{equation}
\sigma
=
\frac{\rho}{
\sqrt{K}\left(1+\sqrt{\frac{\log n_s}{K}}\right)
}
\label{eq:app_tail_sigma_choice}
\end{equation}
ensures that $\|\boldsymbol{\epsilon}\|_2\le\rho$ with probability at least $1-1/\sqrt{n_s}$.

\bibliographystyle{IEEEtran}
\bibliography{ijcai25}

\end{document}